\documentclass[runningheads]{llncs}

\usepackage{listings}
\usepackage[T1]{fontenc}
\usepackage{appendix}
\usepackage{booktabs}

\usepackage{amsmath}
\usepackage{graphicx}

\begin{document}
	
	\title{OpenScout v1.1 mobile robot:\\ a case study on open hardware  continuation}
	
	\author{Bartosz Krawczyk\orcidID{0009-0002-4699-0399}
		\and Ahmed Elbary\orcidID{0009-0001-8159-8627}
		\and Robbie Cato
		\and Jagdish Patil
		\and Kaung Myat
		\and Anyeh Ndi-Tah
		\and \\ Nivetha Sakthivel
		\and Mark Crampton 
		\and \\ Gautham Das\orcidID{0000-0001-5351-9533}
		\and Charles Fox\orcidID{0000-0002-6695-8081}
	}
	
	\authorrunning{B. Krawczyk et al.}
	
	\institute{School of Engineering and Physical Science, University of Lincoln, UK\\}
	
	\maketitle
	
	\begin{abstract}
		
		OpenScout is an Open Source Hardware (OSH) mobile robot for research and industry. It is extended to v1.1 which includes simplified, cheaper and more powerful onboard compute hardware; a simulated ROS2 interface; and a Gazebo simulation. Changes, their rationale, project methodology, and results are reported as an OSH case study. 
		
	\end{abstract}
	
	\section{Introduction}

	\begin{figure}
		\centering
		\includegraphics[width=0.7\linewidth]{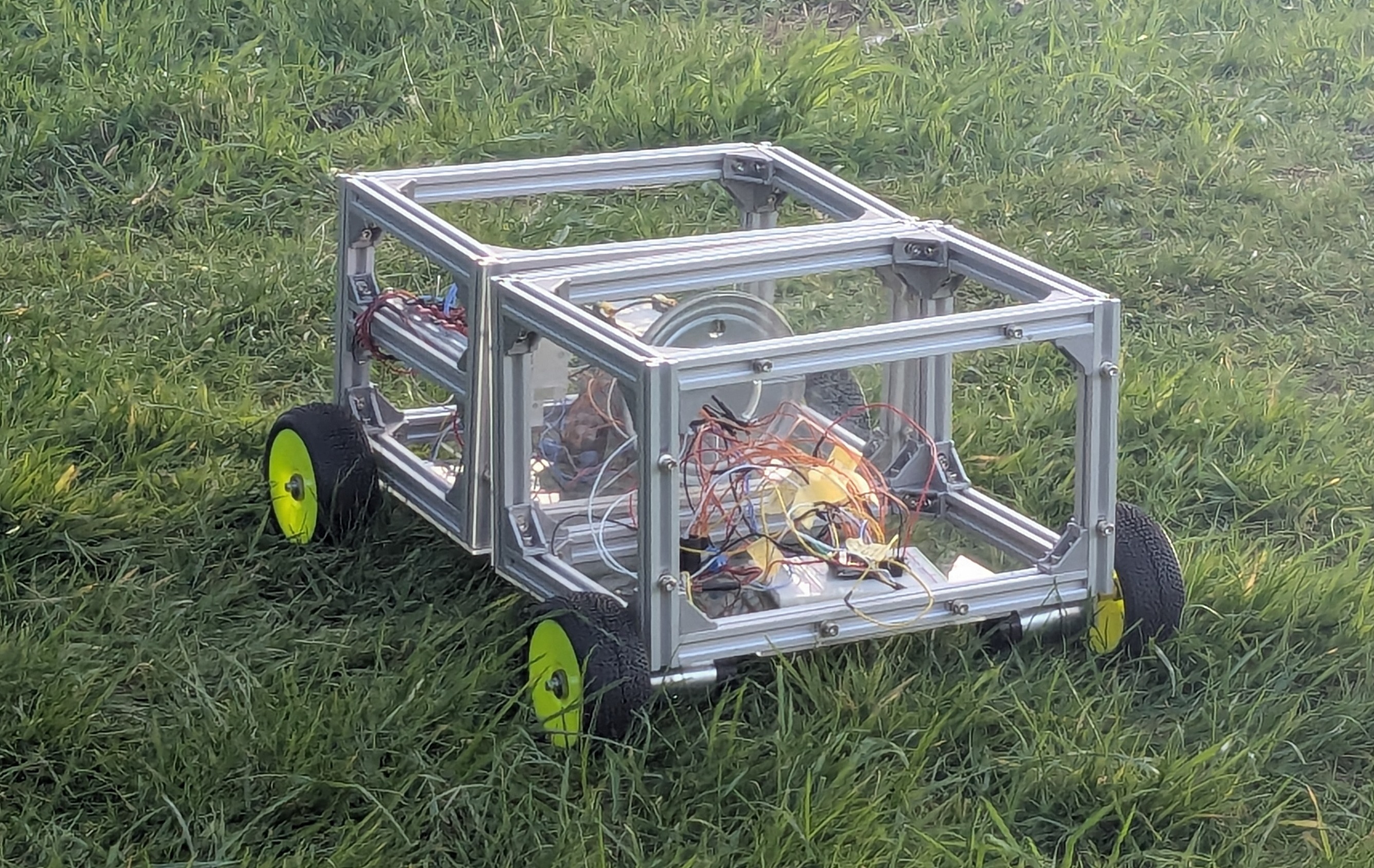}
		\caption{An OpenScout v1.1 driving on a grass field.}
		\label{fig:openscout}
	\end{figure}

	Open Source Hardware (OSH) \cite{ref_osh} is increasingly important in robotics, offering accessible, scientifically reproducible and extendable alternatives to proprietary systems. OpenScout~\cite{ref_article1} is an approximately $0.5m$ by $0.5m$ off-road OSH mobile platform, designed as a base to be modified, experimented with, and replicated to accelerate robotics workflows. It can be safely mounted with $6kg$ of payload, moving at $0.45ms^{-1}$ ($22kg$  dry weight with $0.6ms^{-1}$ linear velocity), with $1h$ of battery life. We consider this to be a `medium size' robot, meaning that it can carry similar loads to a human (with `small' robots not able to do this), but without risk of serious injury to humans on collision (which is what we consider to be a defining trait of `large' robots). 
	
	While OSH lowers barriers to entry, many projects struggle to evolve beyond an initial release, stagnating due to lack of ongoing development, insufficient documentation, or limited uptake.	This paper serves as a case study in maintaining and advancing an OSH project beyond its initial version. By documenting iterative improvements, failures, and rationale, it aims to provide insights that may benefit other robotics initiatives facing similar challenges.
	
	\subsection{Related work}
	
	OSH is the antithesis of conventional commercial philosophy in which a company develops a tool or vehicle, sells builds, and keeps all the design files inaccessible. In OSH, the design itself is released as public source code including CAD, Bill of Materials (BoM), and build instructions, under open licences which allow modifications but require them to be contributed back under the same licence.   Conventional designs become unobtainable and unrepairable when their company owner goes bankrupt or discontinues the product.  OSH enables anyone to pick up the design and continue to work on builds and repairs.  This increases confidence in designs and encourages their adoption.   
	
	Perhaps the most known OSH project is Arduino \cite{ref_arduino}. Other usable OSH projects include robot arms \cite{ref_robotarm}, the Jet Propulsion Laboratory (JPL) OSH rover \cite{ref_osrover} based on a NASA Mars rover, and a tractor \cite{ref_lifetrac} which is designed to be an accessible and maintainable farm `workhorse'.
	
	The first OpenScout\cite{ref_article1} was created in 2021 at the University of Lincoln. Since then, every year a new cohort of MSc students has built and improved upon it.   
	
	Before OpenScout, there was no clearly peer reviewed and fully OSH-licenced standard \cite{ref_article2} for reproducible and verifiable medium sized mobile robotics research. While the large JPL OSH Mars rover \cite{ref_osrover} was designed for research, the 1600 USD BoM and 100 person-hours to manufacture may be a hurdle for widespread adoption. Perhaps the OSH design most comparable to OpenScout is the robot developed by Betancur et al.~\cite{ref_betancur}, although it is far smaller. OpenScout is smaller than the JPL rover, and the main advantage is the 350 USD BoM, 6 to 20 person-hour build time (verified each year), and lower complexity.

	\section{Method}
	
	Since 2021, OpenScout has been modified, extended and forked by a collaboration between robotics academic staff and MSc students. The core model is that each year staff set high-level objectives and provide support, while the current cohort of MSc students create and build new designs as part of their degree. Aside from proving to be a valuable education tool, the OpenScout has undergone significant alteration and updates since the original design~\cite{ref_article1} was created.
	
	This report focuses on the 2025 development cycle. Initial high-level changes were selected by observing trends in other OSH projects, which were then evaluated against OpenScout core objectives (low BoM and build complexity). Other needed changes were flagged whilst mid-way through development, based on discovering and resolving practical limitations.

	\section{Results}
	
	\subsection{Compute update}

	OpenScout v1.1 replaces the Arduino Mega~\cite{ref_arduino} microcontroller (MCU) with  ESP32~\cite{ref_esp32} (ESP32-CH340C Type-C), as it is cheaper, smaller, and more powerful. Previously, the Arduino Mega handled all control tasks, including communication via an external antenna which could only input RF signals from a specific handset controller.  ESP32 includes both WiFi and Bluetooth which provide further options for control, including linking to automation stacks and generic gamer hand controllers.   The ESP32 takes input voltage from the main battery via a 5V buck converter rather than the previously required secondary battery.  These optimisations decrease the part count, base weight, space usage, and cost.
	
	\begin{figure}
		\centering
		\includegraphics[width=0.55\linewidth]{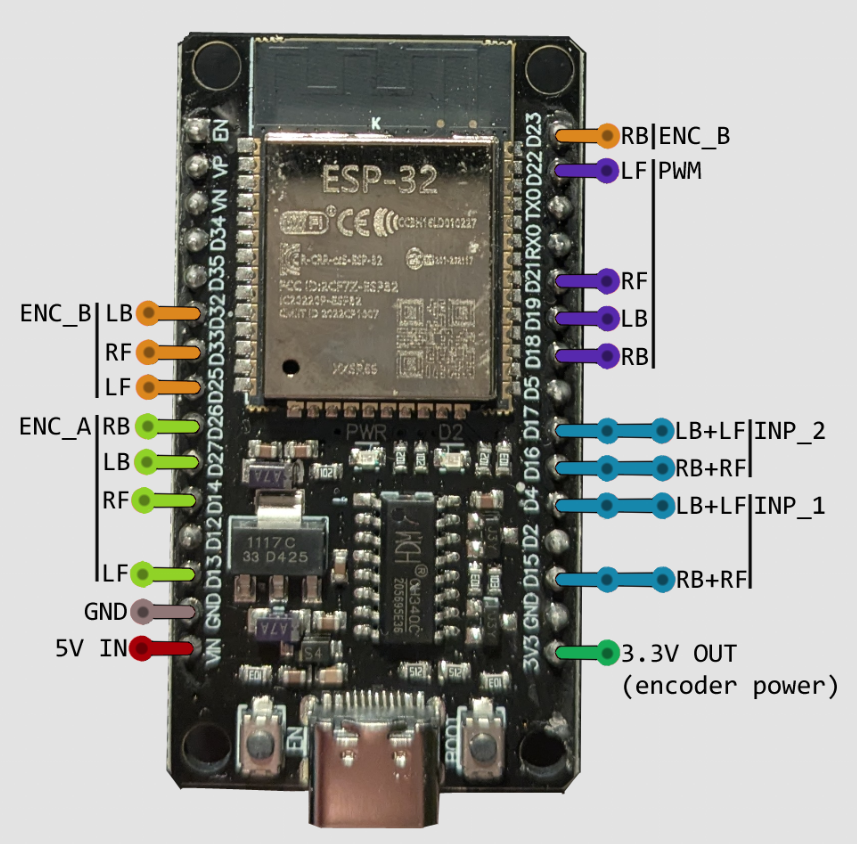}
		\caption[]{ESP32 pinout diagram, showing pin usage for OpenScout v1.1. \textit{LB} is left back motor, \textit{RF} is right front motor, etc.}
		\label{fig:esppinout}
	\end{figure}
	
	While there are differences in internal workings, such as interrupts or handling of digital to analogue signal processing, the ESP32 software is effectively interchangeable with Arduino MCUs.  Additionally, while the pin count, functions and layout are different, it is possible to configure ESP I/O pins to provide the same hardware interface as Arduino. Figure \ref{fig:esppinout} illustrates the exact pins used. Notably certain pins output to two wires. This is possible because the OpenScout is a differential steering platform, thus all left and right motors must act together. Using Hall effect encoders, each motor is independently modulated to match the target velocity. This is a common and efficient drive system. 
	
	\subsection{Wheels failure}
	
	During the verification of the GitHub repository it was found that the closed-source wheels are no longer being produced. This highlights a critical challenge with OSH projects: part availability. Closed source subcomponents are allowed by OHL-CERN~\cite{ref_cernohl} licences, and might seem cost-effective at design time. However, they may then go out of production and be technically and/or legally impossible to reproduce, which adds substantial debt to maintaining the project.
	
	This is a failure of the OpenScout process. Fortunately, thanks to advances in 3D printing, the proprietary part problem has a solution as once a design for a part has been created it can be reprinted forever. A deeper OSH parametric printable wheel design similar to the one created by Robinel~\cite{ref_wheel} could be used, although that particular wheel design is incomplete. Designs of seemingly simple OSH components like this might become an important backbone of OSH eventually, just as there are commonly reused libraries in open source software.
	
	\subsection{Lightweight ROS2 interface via MQTT}
	
	\begin{figure}
		\centering
		\includegraphics[width=0.6\linewidth]{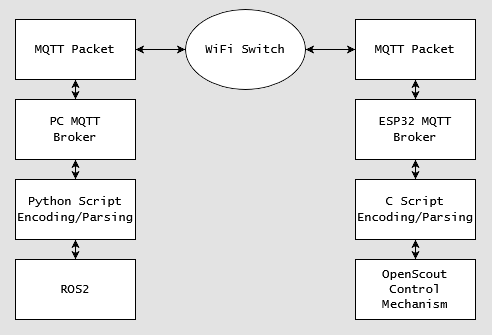}
		\caption{Example PC to OpenScout data pipeline.}
		\label{fig:mqttdiagram}
	\end{figure}
	
	A wireless interface was developed to take advantage of the newly implemented WiFi-enabled MCU. The MQTT (Message Queuing Telemetry Transport) protocol was selected, an open lightweight messaging protocol, which enables efficient wireless communication between the robot and devices such as external PCs running heavyweight AI control systems.
	
	MQTT is based on the publish-subscribe pattern. Devices (clients) publish messages to specific topics, and other devices subscribed to those topics receive the messages. A MQTT broker manages this communication.  OpenScout v1.1 \textit{does not run ROS2}, rather it has a lightweight interface which can read and write a subset of ROS2 message syntax at MQTT level to communicate basic commands. Fig.~\ref{fig:mqttdiagram} illustrates this. A Python tutorial example program is included which demonstrates sending a MQTT message from ROS2 to  ESP32.
	
	This set-up was implemented as a workaround to the 1.3MB of ESP32 internal storage, which is insufficient for ROS2. Micro-ROS could have been used, but it was rejected as it was easier to port the existing MCU program to ESP32. This might increase difficulty of modification in the future, to mitigate this an explanatory note was added. The system can be extended in the future by adding an external antenna to the ESP32 to boost the communication range.
	
	\subsection{Simulation}
	
	\begin{figure}
		\centering
		\includegraphics[width=0.5\linewidth]{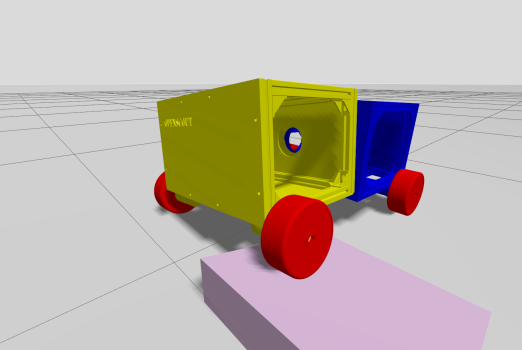}
		\caption{The simulated OpenScout v1.1 on a ramp.}
		\label{fig:Openscout1}
	\end{figure}
	
	We developed a simulation which mirrors the physical platform, including simulating its  ROS2 interface and topic structure.   Several robot simulators were evaluated.  Closed-source options were rejected as they conflict with OpenScout's philosophy. Gazebo~\cite{ref_gazebo} was ultimately chosen due to its ROS2 compatibility and history, including its physics friction simulation capabilities. 
	
	Gazebo Harmonic (latest LTS release) was selected. A major drawback is the lack of Open Dynamics Engine (ODE) support in newer versions of Gazebo, which were previously available in Gazebo Classic. The original plan was to use ODE with Trimble's skid-steer calibration protocol~\cite{ref_trimble} which can be used to calibrate the simulation to millimeter precision given the correct floor material. Unfortunately, this is no longer possible in Gazebo Harmonic. The simulated robot movement is calibrated to $0.5ms^{-1}$ and $0.35rads^{-1}$, which is approximately that of an OpenScout v1.1 with $3kg$ of payload.
	
	CAD files from the original OpenScout project were used to create the robot physical and graphical models. To improve performance, unnecessary elements like screws and bolts were removed, reducing the computational load. The model is defined only in SDF for simplicity. The alternative would be to use URDF or XACRO, which are better integrated into the toolchain. Ultimately, SDF was chosen as any errors resulting from this are insignificant given the ODE issue.
	
	\subsection{Versioning and quality}
	
	Like most OSH projects, OpenScout v1 did not have a concrete versioning plan.   It was peer reviewed and published in a journal and an archival source repository, so is permanently available.   However some of its components have gone out of production or been superseded in capabilities, so updates are, and will continue to be, necessary.   
	
	A continuing challenge has been how to balance update needs with quality control.  Small changes to the peer-reviewed design invalidates its quality status, but are not sufficient to warrant a full re-review and publication.  As with software: many users will prefer older quality assured versions to newer but unassured versions.   A possible strategy is to regularly package {\em sets} of improvements for peer review and publication in venues that match their current quality.   This includes the present v1.1 report, and potentially a future journal v2.0 re-publication once the new features become stable, specified, documented (including rewriting all of the 100+ step build instructions to accommodate them) and quality assured.
	
	As with open source software, managing multiple forks by individuals and teams has been a challenge.  Everyone wants to do OSH design, fewer want to do more useful QA. Until it reaches 2.0 quality, v1.1. remains a fork of the original OpenScout rather than a merge into its main stable design.   At least one other fork of the original has appeared \cite{ref_R4}, which like ours is not yet at 2.0 quality due to documentation and QA needs.   As with software, there will be challenges as multiple forks from stable versions compete, co-exist, or merge.
	
	\section{Conclusion}
	
	OpenScout v1.1~\cite{ref_openscout2} improves the OpenScout OSH mobile robot, including lowering cost and difficulty of hardware build, adding a simulated ROS2 interface, and a simulation. This enhances OpenScout’s usability, accuracy, interoperability, and flexibility.  OpenScout v1.1 has been built but its design is not yet finalised or documented to a v2.0 quality matching the original OpenScout v1, and finding or creating OSH processes to complete and assure such changes remains challenging.   Designs cannot just be peer-reviewed and published once as available components change over time, but full peer review is not possible for the small incremental modifications which are the usual form of contributions in both open hardware and software.   Closed source subcomponents are especially vunerable, so creation and use of deeper open subcomponents -- such as OSH wheels -- would be valuable future contributions.
	
	The release, including a video demo, is available:   \\ \quad \begin{url}     https://github.com/ilovemicroplastics/OpenScoutV1-1\end{url}.


\begin{thebibliography}{8}
		\bibitem{ref_osh}
		Pearce JM. (2012) Building research equipment with free, open-source hardware. 
		\\Science 337.6100:1303-1304
		\\doi:10.1126/science.1228183
		
		\bibitem{ref_article1}
		Carter SJ, Tsagkopoulos NC, Clawson G, Fox C. (2023) OpenScout: Open Source Hardware Mobile Robot.
		\\Journal of Open Hardware 7(1):1-11
		\\doi:10.5334/joh.54
		
		\bibitem{ref_arduino}
		Arduino: Arduino Mega 2560 Rev3. \\https://store.arduino.cc/products/arduino-mega-2560-rev3
		\\Accessed 01 July 2025
		
		\bibitem{ref_robotarm}
		Open Source Robot Arms: Source Robotics.
		\\https://source-robotics.com/
		\\Accessed 01 July 2025
		
		\bibitem{ref_osrover}
		Open Source Rover: Open Source Rover Documentation.  \\https://open-source-rover.readthedocs.io/en/latest/
		\\Accessed 01 July 2025
		
		\bibitem{ref_lifetrac}
		Open Source Ecology: LifeTrac. (2022) \\https://wiki.opensourceecology.org/wiki/LifeTrac
		\\Accessed 01 July 2025
		
		\bibitem{ref_article2}
		Bonvoisin J et al (2020) Standardisation of Practices in Open Source \\Hardware. Journal of Open Hardware 4(1):1-11
		\\doi:10.5334/joh.22 
		
		\bibitem{ref_betancur}
		Betancur-Vásquez D et al (2021) Open source and open hardware mobile robot for developing applications in education and research. \\HardwareX 10(1):e00217
		\\doi:10.1016/j.ohx.2021.e00217
		
		\bibitem{ref_esp32}
		ESPRESSIF: ESP32.
		\\https://www.espressif.com/en/products/socs/esp32
		\\Accessed 01 July 2025
		
		\bibitem{ref_cernohl}
		CERN: CERN Open Hardware License.
		\\https://cern-ohl.web.cern.ch
		\\Accessed 18 July 2025
		
		\bibitem{ref_wheel}
		Robinel A (2016) OpenWheel : parametric OSH wheels/tyres/tracks. hackaday \\https://hackaday.io/project/16024-openwheel-parametric-osh-wheelstyrestracks
		\\Accessed 01 July 2025
		
		\bibitem{ref_gazebo}
		Gazebo: Gazebo Official Website.
		\\https://gazebosim.org/
		\\Accessed 01 July 2025
		
		\bibitem{ref_trimble}
		Russel R, Fox C (2023) Skid-steer friction calibration protocol for digital twin creation.
		\\In:TAROS2023
		
		\bibitem{ref_R4}
		Waltham C, Perrett A, Soni R, Fox C (2025) R4: rapid reproducible robotics research open hardware control system.
		\\Journal of Open Hardware 9(1):1-11
		\\doi:10.5206/joh.v9i1.22878
		
		\bibitem{ref_openscout2}
		Krawczyk B, Elbary E, Cato R et al (2025) OpenScoutV1.1. \\https://github.com/ilovemicroplastics/OpenScoutV1-1
		\\Accessed 18 July 2025
	\end{thebibliography}
\end{document}